\documentclass[a4paper,UKenglish,cleveref, autoref, thm-restate]{oasics-v2021}
\usepackage{booktabs}
\usepackage{amsmath}
\usepackage{float}
\usepackage[T1]{fontenc}
\usepackage{mathtools}
\usepackage[thinc]{esdiff}
\usepackage{algorithm}
\usepackage{algpseudocode}
\DeclareMathOperator*{\argmax}{argmax}

\title{ Claim Verification using a Multi-GAN based Model} 

\titlerunning{Claim Verification using a Multi-GAN based Model} 

\author{Amartya Hatua}{University of Houston, 4800 Calhoun Rd, Houston, TX 77004  }{ahatua@central.uh.edu}{}{}

\author{Arjun Mukherjee}{University of Houston, 4800 Calhoun Rd, Houston, TX 77004  }{arjun@uh.edu}{}{}

\author{Rakesh M. Verma}{University of Houston, 4800 Calhoun Rd, Houston, TX 77004  }{rmverma@cs.uh.edu}{}{}



\category{} 

\relatedversion{} 
\authorrunning{Amartya Hatua, Arjun Mukherjee, Rakesh M. Verma}

\acknowledgements{Research was supported in part by grants NSF 1838147, NSF 1433817 and ARO W911NF-20-1-0254. Verma is the founder of Everest Cyber Security and Analytics, Inc. The views and conclusions contained in this document are those of the authors and not of
the sponsors. The U.S. Government is authorized to reproduce and distribute reprints for Government purposes notwithstanding any copyright notation herein.}

\nolinenumbers 

\EventEditors{John Q. Open and Joan R. Access}
\EventNoEds{2}
\EventLongTitle{42nd Conference on Very Important Topics (CVIT 2016)}
\EventShortTitle{CVIT 2016}
\EventAcronym{CVIT}
\EventYear{2016}
\EventDate{December 24--27, 2016}
\EventLocation{Little Whinging, United Kingdom}
\EventLogo{}
\SeriesVolume{42}
\ArticleNo{23}

\begin{document}

\maketitle

\begin{abstract}

This article describes research on  claim verification carried out using a multiple GAN-based model. The proposed model consists of three pairs of generators and discriminators. The generator and discriminator pairs are responsible for generating synthetic data for supported and refuted claims and claim labels. A theoretical discussion about the proposed model is provided to validate the equilibrium state of the model. The proposed model is applied to the FEVER dataset, and a pre-trained language model is used for the input text data. The synthetically generated data helps to gain information which helps the model to perform better than state of the art models and other standard classifiers.

\end{abstract}

\section{Introduction}
\label{sec:typesetting-summary}

Misleading claims and news are prevalent phenomena in our day-to-day life. Sometimes these are extremely difficult to identify, and as a result, they are causing serious problems in our lives. This problem makes the research on  claim verification extremely important and necessary. Fake news can be broadly classified into three categories \cite{rubin2015deception}: i) Serious fabrications (uncovered in mainstream or participant media, yellow press or tabloids); ii) Large-scale hoaxes; and iii) Humorous fakes (news satire, parody, game shows). To solve this problem, the research on this subject evolved from the use of knowledge-base oriented methods to sophisticated deep learning-based techniques. 

In \cite{mihalcea2009lie} Mihalcea, and Strapparvva used natural language processing (NLP) techniques to detect fake news, which is considered one of the earliest attempts to solve this problem using NLP. Mihalcea and Strapparvva used tokenization and stemming for preprocessing the data and applied Naıve Bayes algorithm and Support Vector Machine (SVM) for the classification. In recent research the linguistic style \cite{1}, \cite{2}, \cite{3} and source of the text are considered as the most critical factors to decide the genuineness of a fact or claim.

 Other than that, sometimes multiple sources of particular claims are used as external resources for  claim verification. In \cite{1}, Hannah et al.  compared the linguistic characteristics of real news with satire, hoaxes, and propaganda. In this research, they presented a case study based on the data collected by PolitiFact.com, where they used Glove for the embedding of the test data, and Long Short Term Memory (LSTM) for the prediction. To improve their result, they concatenate the  Linguistic Inquiry and Word Count (LIWC) features \cite{pennebaker2001linguistic} with LSTM output vectors before undergoing the activation layer. LIWC features played a very vital role in  claim verification research. LWIC extracts essential words in the text that are part of psycho-linguistic categories and helps in content analysis \cite{krippendorff2018content, neuendorf2015content}. This research work has been further extended by Kashyap et al.  \cite{popat2018declare}, and they proposed an End-to-end Framework for Credibility Analysis. The proposed end-to-end framework is capable of aggregating information from external evidence articles, the language of these articles, and the trustworthiness of their sources. It also helps in generating informative features for user-comprehensible explanations \cite{popat2018declare}. The use of external information sources is a very effective technique for  claim verification research, like Ravali et al. in \cite{7}, \cite{pasternack2011making}, \cite{ge2013multi}, \cite{li2014resolving}, \cite{wan2016truth} used external sources for similar types of tasks. Ravali et al. proposed a novel method based on correlations between different sources of news in \cite{7}. To find the correlation between sources, joint precision and joint recall are used. 

Jeff Pasternack et al. introduced a generalized fact-finding framework in \cite{pasternack2011making}, in which they tried to solve a fact finding problem while different authors are making conflicting claims. Similarly, \cite{ge2013multi}, \cite{li2014resolving}, \cite{wan2016truth}
 also used inconsistent sources and information to verify the facts and claims. Liang Ge et al. \cite{ge2013multi} proposed a two-step procedure that calculates the degree of information consistency and identifies the underlying common reason for the inconsistency, and calculates a consistent score for each item. Likewise, Q. Li et al. \cite{li2014resolving} proposed an optimization framework in which truths and reliable sources are considered as two sets of unknown variables, and the framework aims to minimize the deviation between the truths and the multi-source observations. A generalized algorithm called TruthFinder is proposed in \cite{wan2016truth}, which utilizes the information of different related websites to perform fact-checking. In most recent research on this topic, applications of deep learning techniques are very prevalent. The fake news detection research by Anshika et al.  \cite{choudhary2020linguistic}, proposed a sequential neural model which helps to identify syntactic, grammatical, sentimental, and readability features of certain news. Using deep learning techniques, Yang Yang et al. \cite{yang2018ti} proposed text and Image information based Convolution Neural Network (TI-CNN), which uses both text and images as evidence for fact-checking. In this model, CNN is used for feature extraction from both text and images.

Recently, FEVER dataset has gained a lot of traction in the researcher community \cite{thorne2018fact}, 
\cite{thorne2019adversarial}, \cite{thorne2019fever2}, hence for our  claim verification research we are using FEVER dataset. In earlier research on the FEVER dataset, most of the researchers followed a pipeline suggested by the baseline model \cite{fever}. The pipeline consists of three phases in a sequence. These phases are identifying relevant wiki articles, extracting the appropriate supporting sentences, and determining the truthfulness of the claim. Most of the earlier researchers implemented the wiki article identifying  phase by Wikipedia API, token matching techniques and the AllenNLP framework  \cite{Gardner2017AllenNLP}.  For sentence selection most of the earlier researchers used TF-IDF based method, sequence matching neural network, and some ranking based methods. The classification task is done using a TF-IDF based approach in the base model, while Neural network based models, different natural language inference models, and deep learning based models are also used later.

\begin{figure} [h]
\centering
\includegraphics[scale=0.70]{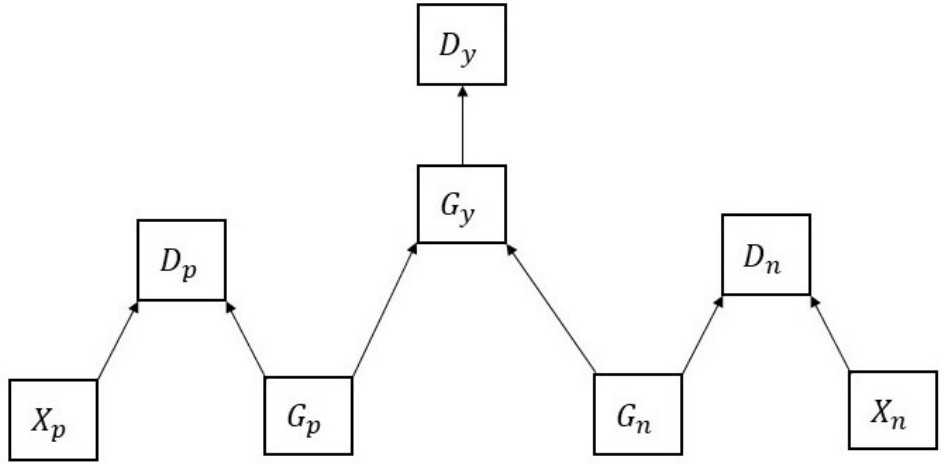}
\caption{Schematic diagram of proposed model}
\label{fig1}
\end{figure}

In this research, we proposed a GAN \cite{gan} based model to perform the claim verification job. The proposed model is  inspired by two GAN based Positive Unlabeled (PU) learning models proposed by Ming et al.  \cite{hou2017generative} (GenPU) and Yang et al. \cite{fan}. The proposed GAN based model consists of three pairs of generators and discriminators. These generator and discriminator pairs are responsible for generating positive or supported claims, negative or refuted claims, and class labels of the claims from a global perspective. Fig. \ref{fig1} shows the proposed model.

This model uses three generators $(G_p, G_n, G_y)$ and three discriminators $(D_p, D_n, D_y)$. $G_p$ is responsible for generating positive claims and $Dp$ discriminates between original positive claims and synthetically generated positive claims. $G_n$ and $D_n$ are responsible for similar functions for negative claims.  $Gy$ and $Dy$ get the input from the data generated by $G_p$ and $G_n$ and generate a class label (0/1) and the $D_y$ is the discriminator for $G_y$.

\section{Proposed Methodology}

In the proposed methodology three GAN units are used. These three units are responsible for generating positive samples Equation \ref{Eq:1}, negative samples Equation \ref{Eq:2} and class labels Equation \ref{Eq:3}. Algorithm \ref{alg:arraygan} gives the steps to train the generators and discriminators. 

\begin{equation}
\label{Eq:1}
\min_{G_p}\max_{D_p} V(D,G) = \mathop{\mathbb{E}_{x \sim p_p{(x)}}} log(D_p(x))+\mathop{\mathbb{E}_{z \sim p_z{(z)}}} log(1-D_p(G_p(z)))
\end{equation} 

\begin{equation}
\label{Eq:2}
\min_{G_n}\max_{D_n} V(D,G) = \mathop{\mathbb{E}_{x \sim p_n{(x)}}} log(D_n(x))+\mathop{\mathbb{E}_{z \sim p_z{(z)}}} log(1-D_n(G_n(z)))
\end{equation} 

\begin{equation}
\label{Eq:3}
\begin{split}
\min_{G_p,G_n,G_y}\max_{D_y} V(D,G) = \mathop{\mathbb{E}_{x \sim p{(x)}}} log(D_y(x))+ 
\pi_p\mathop{\mathbb{E}_{z \sim p_z{(z)}}} log(1-D_y(G_y(G_p(z))))+
\\ \pi_n\mathop{\mathbb{E}_{z \sim p_z{(z)}}} log(1-D_y(G_y(G_n(z))))
\end{split}
\end{equation}

\begin{algorithm}[h]
\caption{Algorithm}
\label{alg:arraygan}
\begin{algorithmic}[1]
\Statex
\For {training iterations} 
    \State {\# \textbf{update} discriminator networks {$D_p, D_n, D_y$} \#}
    
    \State {sample mini-batch of noise examples $\{z^i\}^{m}_{i=1}$ 
    from noise prior $p_z(z)$}
    
    \State {sample mini-batch of positive examples $\{x_p\}^{m}_{i=1}$ 
    from noise prior $p_p(x)$}
    
    \State {sample mini-batch of negative examples $\{x_n\}^{m}_{i=1}$ 
    from noise prior $p_n(x)$}
    
    \State {sample mini-batch of examples $\{x\}^{m}_{i=1}$ 
    from noise prior $p(x)$}
    
    \State {update the positive discriminator $D_p$ by ascending its
stochastic gradient:$\nabla_{\theta_{D_p}} \frac{1}{m} \sum_{i=1}^{m} \pi_p [log(D_p(x^i_p)) + log(1-D_p(G_p(z^i)))]$}

    \State {update the negative discriminator $D_n$ by ascending its
    stochastic gradient:$\nabla_{\theta_{D_n}} \frac{1}{m} \sum_{i=1}^{m} \pi_n [log(D_n(x^i_n)) + log(1-D_n(G_n(z^i)))]$}

    \State {update the discriminator $D_y$ by ascending its
    stochastic gradient:$\nabla_{\theta_{D_y}} \frac{1}{m} \sum_{i=1}^{m} \pi_p [log(D_y(x^i)) +  \pi_p log(1-D_y(G_y(G_p(z))) + \pi_n log(1-D_y(G_y(G_n(z)))]$}
    
    \State {\# \textbf{update} generator networks {$G_p, G_n, G_y$} \#}
    
    \State {sample mini-batch of noise examples $\{z^i\}^m_{i=1}$from noise
prior $p(z)$}

    \State {update the positive generator $G_p$ by descending its stochastic
gradient: $\nabla_{\theta_{G_p}} \frac{1}{m} \sum_{i=1}^{m}  \pi_p [-log(D_p(G_p(z^i)) - log(D_y(G_p(z^i))]$}
    
    \State {update the negative generator $G_p$ by descending its stochastic
gradient: $\nabla_{\theta_{G_n}} \frac{1}{m} \sum_{i=1}^{m}  \pi_n [-log(D_n(G_n(z^i)) - log(D_y(G_n(z^i))]$}

    \State {update the class label generator $G_y$ by descending its stochastic
    gradient: $\nabla_{\theta_{G_y}} \frac{1}{m} \sum_{i=1}^{m}  [- \pi_p log(D_y(G_p(z^i)) - \pi_n log(D_y(G_n(z^i))]$}
    
\EndFor

\State return $G_y$

\end{algorithmic}
\end{algorithm}

The proposed model can handle only supported and refuted claims. 
$D_y$ will be trained with both supported and refuted claims, while $D_p$ and $D_n$ will be trained with only supported and  refuted claims separately hence, $D_y$ is a more powerful discriminator compared to $D_p$ and $D_n$. There is a possibility that $D_p$ or $D_n$ will assign some sentences generated by $G_p$ and $G_n$ wrongly. As $D_y$ has the global view of both supported and refuted claims, it is capable of classifying them. Consider a situation: $G_p$ generates $Y_p$ (a synthetic positive claim). In the next step, $Y_p$ is the input to $G_y$, and $G_y$ is generating 1 (positive class label). The output of $G_y$ and input of $G_p$ is the input to the discriminator state ($D_y$). If $D_y$ classifies the $Y_p$ as positive, then there is no penalty that will be added to $G_y$ and $G_p$ otherwise penalties will be added to both $G_p$ and $G_y$. Consider another situation, where the $G_y$ generates 0 (negative class label) for an input of $Y_p$ and $D_y$ is also classifies the $Y_p$ as negative, then a penalty will be added to $G_p$, not $G_y$. So $D_y$ is acting as a global discriminator. Equation \ref{Eq:4} is the loss function for the generator $G_y$. $\pi_p$ and $\pi_n$ are the probabilities of positive and negative claims in the dataset.

\begin{equation}
\label{Eq:4}
\begin{split}
    L(y) = \pi_p [D_y(G_p(z)) log(D_y(G_y(G_p(z)))) + (1-D_y(G_p(z))) log(1-D_y(G_p(z)))] + \\
    \pi_n [D_y(G_n(z)) log(D_y(G_y(G_n(z)))) + (1-D_y(G_n(z))) log(1-D_y(G_n(z)))]
\end{split}
\end{equation}

For a GAN system achieving  the equilibrium condition is very important. In the present context, to find the equilibrium condition, first, we need to find the optimal conditions for discriminators. Using the optimal conditions of the discriminators, the minimization conditions for the generator can be obtained. Considering the generators ($G_p$, $G_n$, $G_y$) are fixed, and $\pi_p$ and $\pi_n$ are the probabilities of positive and negative claims in the dataset. So in the in equilibrium condition the distribution of positive generated data ($p_{gp}(x)$) and negative generated data ($p_{gn}(x)$) will follow the below mentioned Equations \ref{Eq:5} and \ref{Eq:6}. In Equations \ref{Eq:5} and \ref{Eq:6}, $p_p(x)$ and $p_n(x)$ are the positive and negative class probability distributions. 

\begin{equation}
\label{Eq:5}
p_{gp}(x) = p_p(x)
\end{equation} 
\begin{equation}
\label{Eq:6}
p_{gn}(x) = p_n(x)
\end{equation}

The optimal discriminator functions $D_p^{*}(x)$, $D_n^{*}(x)$, $D_y^{*}(x)$ can be derived by differentiate Equation \ref{Eq:1}, \ref{Eq:2} and \ref{Eq:3}.

\begin{equation}
\label{Eq:11}
    D^{*}_P(x) = \frac{p_p(x)}{p_p(x)+p_{gp}(x)}
\end{equation}

\begin{equation}
\label{Eq:12}
    D^{*}_n(x) = \frac{p_n(x)}{p_n(x)+p_{gn}(x)}
\end{equation}

\begin{equation}
\label{Eq:16}
\begin{split}
    \min_{G_p,G_n,G_y}\max_{D_y} V(D^*,G) = log\left(\frac{p(x)}{p(x)+\pi_pp_{gp}(x)+\pi_np_{gn}(x)}\right) + \\ \pi_plog\left(\frac{\pi_pp_{gp}(x)+\pi_np_{gn}(x)}{p(x)+\pi_pp_{gp}(x)+\pi_np_{gn}(x)}\right) + \\
    \pi_nlog\left(\frac{\pi_pp_{gp}(x)+\pi_np_{gn}(x)}{p(x)+\pi_pp_{gp}(x)+\pi_np_{gn}(x)}\right)
\end{split}
\end{equation}

Using Jensen–Shannon divergence (JSD) \cite{fuglede2004jensen}, we can show that the minimum value of the generators can be achieved when following conditions will be satisfied: 

\begin{equation}
\label{Eq:17}
p_p(x) = p_{gp}(x)    
\end{equation}

\begin{equation}
\label{Eq:18}
p_n(x) = p_{gn}(x)    
\end{equation}

\begin{equation}
\label{Eq:19}
p_y(x) = \pi_pp_{gp}(x)+\pi_np_{gn}(x)
\end{equation}

The derivation steps of the above mentioned equations is presented in Appendix \ref{Appendix:A}.

\section{Data}

FEVER dataset is used for this research. It is a publicly available dataset for claim verification. There are three types of claims present in the dataset i) supported, ii) refuted, iii) information not enough (INE). For every supported and refuted claim there is one or multiple supporting evidence, while for the INE class there is no evidence. All evidence provided in the FEVER dataset is collected from Wikipedia. In most cases, the first few lines of a particular Wikipedia page are taken in FEVER dataset as the evidence.
In Table 1 two examples of the claim, evidence and class label are presented.

\begin{table*}[h]
  \caption{Examples of claim verification}
  \label{tab:table1}
  \begin{tabular}{l}
    \toprule
    \toprule
    \textbf{Claim:} Tetris has sold millions of physical copies.\\
\textbf{Evidence:} It was announced that Tetris has sold more than 170 million\\
copies, approximately 70 physical copies and ...\\
\textbf{Label:} True
    \\
    \toprule
    \textbf{Claim:} Andy Roddick lost 5 Master Series between 2002 and 2010.\\
\textbf{Evidence:}  Roddick was ranked in the top 10 for nine consecutive years \\ between 2002 and 2010, and \\ won five  Masters Series in that period.\\
\textbf{Label:} False
    \\
    \toprule
    \toprule
\end{tabular}
\end{table*}

FEVER training dataset has 80,035 Supported claims, 29,775 Refuted claims, and 35,639 NotEnoughInfo claims. The FEVER 1.0 validation set and test set have 3,333 Support claims, 3,333 Refute claims, and 3,333 NotEnoughInfo claims respectively. FEVER 2.0 has 391 Support claims, 396 Refute claims, and 387 NotEnoughInfo claims respectively. For the experiments we used only supported and refuted claims.

\section{Experiments}

The algorithm described in the previous section is implemented and tested using the FEVER 1.0 and FEVER 2.0 datasets. The steps of the experiments are described in this section. 
\\
\subsection{Data preprocessing:} For this experiment, only `Supported' and `Refuted' claims are considered from the training dataset. In the training dataset, every claim has one or multiple evidence. For a particular claim, its corresponding evidence is concatenated separately. For example, there is a data point with the following claim $(C)$ evidence $(E)$ and label $(L): [C,E <e_1, e_2, e_3>, L].$ The input data format for the further processes will be: $ x = [<C; e_1, L>, <C; e_2, L>, <C; e_3, L>]$. This preprocessed claim evidence pair is used for further experiments. 
\\
\subsection{GAN Implementation:} The implementation of GAN is the central part of this research. There are two types of GAN implemented: text generating GAN and binary class label generating GAN. The text generating GAN is generating synthetic text data for supported and refuted claims. The binary class label generating GAN generates the binary class label for each of the generated claims. To implement text generating GAN, we followed LaTextGAN \cite{donahue2018adversarial}. LaTextGAN follows two phases for the implementation. During the first phase, it creates an encoded space, and in the second phase, it follows the traditional GAN \cite{gan} implementation steps and generates synthetic data in the encoded space. Finally, the synthetically generated data is decoded into normal text data. On the other hand, the implementation of binary labels generating GAN is similar to the implementation of the traditional GAN \cite{gan}. 
\\
\subsection{GenPU Based Methods:}

The proposed model is inspired by the GenPU. To explore further we have modified GenPU in two variants such as Inverted GenPU and Symmetric GenPU. In case of Inverted GenPU the value functions for the positive and negative text generating GAN are exchanged. Hence the respective value functions become the equations mentioned in Equation 13, 14 and 15.

\begin{equation}
D^*_n = \argmax_{D_n} \mathop{\mathbb{E}_{x \sim p_p{(x)}}} log(D_n(x)) + \mathop{\mathbb{E}_{z \sim p_z{(z)}}}log(1-D_u(G_n(z)))
\end{equation}

\begin{equation}
\min_{G_p}\max_{D_p} V(D,G) = -\mathop{\mathbb{E}_{x \sim p_p{(x)}}} log(D^*_n(x))-\mathop{\mathbb{E}_{z \sim p_z{(z)}}} log(1-D^*_n(G_n(z)))
\end{equation} 

\begin{equation}
\min_{G_n}\max_{D_n} V(D,G) = \mathop{\mathbb{E}_{x \sim p_p{(x)}}} log(D_p(x))+\mathop{\mathbb{E}_{z \sim p_z{(z)}}} log(1-D_p(G_p(z)))
\end{equation} 

In Symmetric GenPU the equations for both the value functions are same. The value functions for Symmetric GenPU   are presented in Equation 16 and 17.

\begin{equation}
\min_{G_p}\max_{D_p} V(D,G) = \mathop{\mathbb{E}_{x \sim p_p{(x)}}} log(D_p(x))+\mathop{\mathbb{E}_{z \sim p_z{(z)}}} log(1-D_p(G_p(z)))
\end{equation} 

\begin{equation}
\min_{G_n}\max_{D_n} V(D,G) = \mathop{\mathbb{E}_{x \sim p_p{(x)}}} log(D_p(x))+\mathop{\mathbb{E}_{z \sim p_z{(z)}}} log(1-D_p(G_p(z)))
\end{equation}

\subsection{Other methods:} The performance of the proposed method is compared with other GAN based methods and classifiers. The GAN based models generate synthetic data and the synthetically generated data is added to the original dataset and it helps to create an extended feature space of the FEVER dataset and gives leverage to new features. This synthetically generated data is further classified using positive-unlabeled (PU) learning which considers supported facts as positive class and are added to the existing training dataset. Finally, this extended dataset is used for the training process. The synthetic data is generated using LeakGAN \cite{leakgan} and LaTextGAN \cite{donahue2018adversarial} separately and two different sets of results are collected to compare the performance. 
Other than GAN based methods different deep learning and machine learning based  classification methods are used such as: BERT based classifier \cite{bert}, Graph Convolution Network (GCN) \cite{scarselli2008graph}, Long Short Term Memory (LSTM) \cite{lstm}, Convolution Neural Network (CNN) \cite{cnn}, Support Vector Machine (SVM) \cite{svm}, Naive Bayes \cite{naive}, Random forest \cite{random}, Stochastic Gradient Descent (SGD) \cite{friedman2002stochastic} are also implemented for the claim verification  task. To implement BERT based classifier Huggingface BERT \cite{bert} pretrained transformer is used as tokenizer for the training, validation and testing dataset. The vocabulary size of the pretrained model is 30522 and the size of the hidden layer is 768. Later the pretuned model is fine tuned to classify the claims. In GCN, the point wise mutual information between words is calculated to generate the graph. To implement the CNN five kernels of sizes 2, 3, 4, 5 and 6 are used. For LSTM the input data is encoded using GloVe \cite{pennington2014glove}. The learning rate and batch size for GCN, CCN and LSTM are 0.001, 64 respectively. The Random forest is equipped with 1000 trees and entropy is used as supported criteria for the information gain. The SGB model utilizes hinge loss and L2 penalty. The deep learning models are implemented using PyTorch \cite{NEURIPS2019_9015}, and the Scikit learn library \cite{scikit-learn} is used for machine learning models. 

\section{Results}
In this section the results of the previously discussed experiments will be discussed. All models are trained with the FEVER training dataset and tested with FEVER 1.0 and FEVER 2.0 test dataset. In the Table \ref{tab:table2} and Table \ref{tab:table3} detailed results for each of the models are presented. Each of the experiments is repeated five times. The result for FEVER 1.0 is also compared with previous research work by Yang et al. \cite{fan}. 

\begin{table}[h]
\caption{Result of FEVER 1.0 }
\label{tab:table2}
\begin{tabular}{|l|l|l|l|}
\hline
 & \multicolumn{3}{c|}{FEVER 1.0 Dataset}  \\ \hline
Classifiers                         & Precision    & Recall    & F1 Score\\ \hline
BERT Classifier                     & 0.45  $\pm$ 0.011       & 0.44   $\pm$ 0.010   & 0.44 $\pm$ 0.009          \\ \hline
Leak GAN Based Classifier & 0.65  $\pm$ 0.003       & 0.64  $\pm$ 0.006    & 0.63 $\pm$ 0.003 \\ \hline
LaTextGAN Based Classifier & 0.41   $\pm$ 0.008      & 0.36  $\pm$ 0.016    & 0.30   $\pm$ 0.009            \\ \hline
Graph Convolutional Network & 0.45   $\pm$ 0.015      & 0.44  $\pm$ 0.013    & 0.44  $\pm$ 0.013              \\ \hline
SVM                                 & 0.53  $\pm$ 0.013       & 0.42  $\pm$ 0.013    & 0.38    $\pm$ 0.013            \\ \hline
Naive Bayes                         & 0.41   $\pm$ 0.016         & 0.34   $\pm$ 0.014     & 0.24  $\pm$ 0.015              \\ \hline
Random forest                       & 0.33   $\pm$ 0.011          & 0.33  $\pm$ 0.010        & 0.28    $\pm$ 0.011         \\ \hline
SGD                       & 0.31    $\pm$ 0.023        & 0.22   $\pm$ 0.022       & 0.27    $\pm$ 0.023             \\ \hline
LSTM                  & 0.45  $\pm$ 0.003        & 0.42   $\pm$ 0.004      & 0.004    $\pm$ 0.004                            \\ \hline
CNN                                 & 0.46 $\pm$ 0.012          & 0.44  $\pm$ 0.011       & 0.43    $\pm$ 0.012                 \\ \hline

Inverted GenPU & 0.52 $\pm$ 0.013  & 0.71 $\pm$ 0.023 & 0.60 $\pm$ 0.018
\\ \hline
Symmetric GenPU & 0.33 $\pm$ 0.015 & 0.54 $\pm$ 0.02 & 0.40 $\pm$ 0.016
\\ \hline
Proposed Method                       & 0.50 $\pm$ 0.016        & 0.93  $\pm$ 0.018    & 0.65 $\pm$ 0.018 \\ \hline
Yang et al. result                        & 0.61         & 0.58      & 0.60    \\ \hline
\end{tabular}
\end{table}

In Table \ref{tab:table2} and Table \ref{tab:table3} it can be observed that the F1 score for the proposed method is better than the rest of the models and the previous research. 

The proposed GAN model has three generator-discriminator pairs and each of them is having one loss function. To generate very good quality synthetic data, the loss should be minimized. While training the model we observed losses of three generator-discriminator pairs such as positive loss, negative loss and binary label loss. In the Fig. \ref{fig4}, Fig. \ref{fig5} and Fig. \ref{fig6} it can be observed that the loss of three generator-discriminator pairs is gradually reduced. 

\begin{table}[h]
\caption{Result of FEVER 2.0 }
\label{tab:table3}
\begin{tabular}{|l|l|l|l|}
\hline
 & \multicolumn{3}{c|}{FEVER 2.0 Dataset}  \\ \hline
Classifiers                         & Precision    & Recall    & F1 Score\\ \hline
BERT Classifier                     & 0.46  $\pm$ 0.013          & 0.44    $\pm$ 0.014     & 0.44 $\pm$ 0.013 \\ \hline
Leak GAN Based Classifier & 0.52  $\pm$ 0.023         & 0.51    $\pm$ 0.019     & 0.51   $\pm$ 0.021        \\ \hline
LaTextGAN Based Classifier & 0.42    $\pm$ 0.02        & 0.39  $\pm$ 0.019       & 0.39  $\pm$ 0.019       \\ \hline
Graph Convolutional Network & 0.43 $\pm$ 0.023          & 0.39  $\pm$ 0.013       & 0.37   $\pm$ 0.016        \\ \hline
SVM                                 & 0.40   $\pm$ 0.019           & 0.37  $\pm$ 0.022         & 0.35     $\pm$ 0.019      \\ \hline
Naive Bayes                         & 0.33  $\pm$ 0.030            & 0.22  $\pm$ 0.023       & 0.27 $\pm$ 0.025          \\ \hline
Random forest                       & 0.33   $\pm$ 0.014          & 0.26  $\pm$ 0.017       & 0.29  $\pm$ 0.015       \\ \hline

SGD                       & 0.30 $\pm$ 0.025          & 0.22  $\pm$ 0.029       & 0.26 $\pm$ 0.027      \\ \hline

LSTM     & 0.43   $\pm$ 0.028   & 0.40   $\pm$ 0.039      & 0.39  $\pm$ 0.032                \\ \hline

CNN                                  & 0.41    $\pm$ 0.021         & 0.38   $\pm$ 0.011      & 0.37    $\pm$ 0.018       \\ \hline
Inverted GenPU & 0.58 $\pm$ 0.024 & 0.71 $\pm$ 0.022 & 0.63 $\pm$ 0.012
\\ \hline
Symmetric GenPU & 0.41 $\pm$ 0.016 & 0.55 $\pm$ 0.011 & 0.49 $\pm$ 0.013
\\ \hline
Proposed Method                       & 0.49   $\pm$ 0.061      & 0.97 $\pm$ 0.041     & 0.65 $\pm$ 0.051 \\ \hline

\end{tabular}
\end{table}

\begin{figure}[t]
  \centering
  \begin{minipage}[b]{0.48\textwidth}
    \includegraphics[width=7cm,height=10cm,keepaspectratio]{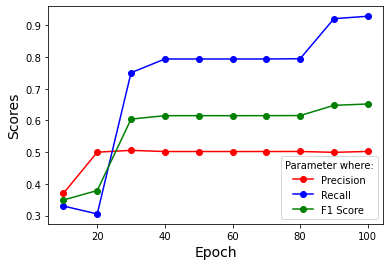}
    \caption{Precision, Recall and F1 Score for FEVER 1.0 Dataset}
    \label{fig2}
  \end{minipage}
  \hfill
  \begin{minipage}[b]{0.48\textwidth}
    \includegraphics[width=7cm,height=10cm,keepaspectratio]{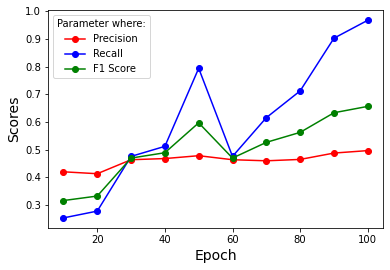}
    \caption{Precision, Recall and F1 Score for FEVER 2.0 Dataset}
    \label{fig3}
  \end{minipage}
\end{figure}

\begin{figure}[H]
  \centering
  \begin{minipage}[b]{0.48\textwidth}
    \includegraphics[width=5cm, height=4cm]{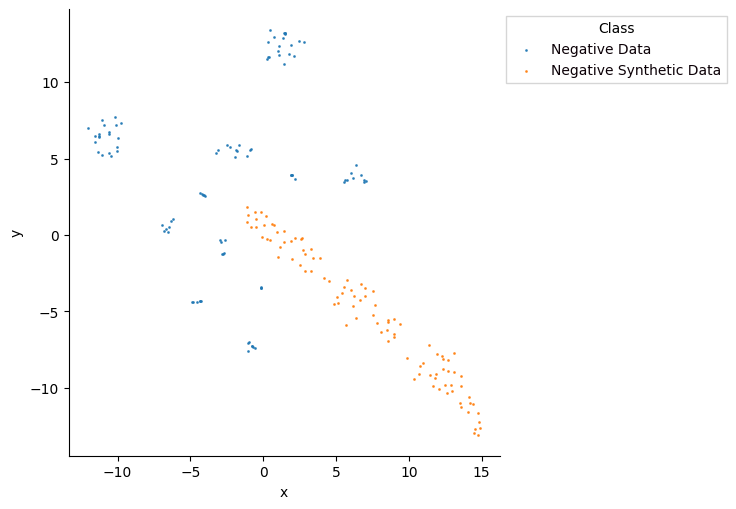}
    \subcaption{Epoch = 25}
  \end{minipage}
  \hfill
  \begin{minipage}[b]{0.48\textwidth}
    \includegraphics[width=5cm, height=4cm]{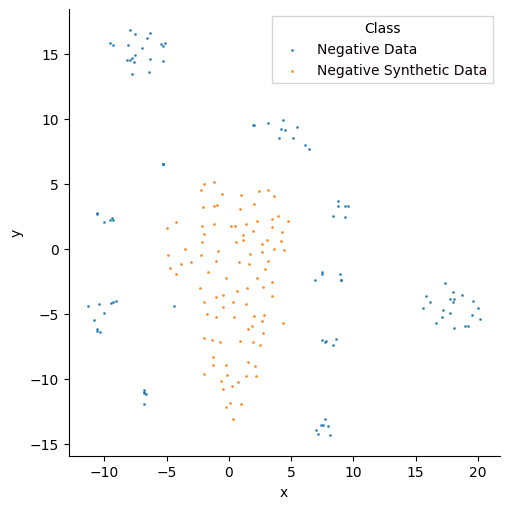}
    \subcaption{Epoch = 50}
  \end{minipage}
    \begin{minipage}[b]{0.48\textwidth}
    \includegraphics[width=5cm, height=4cm]{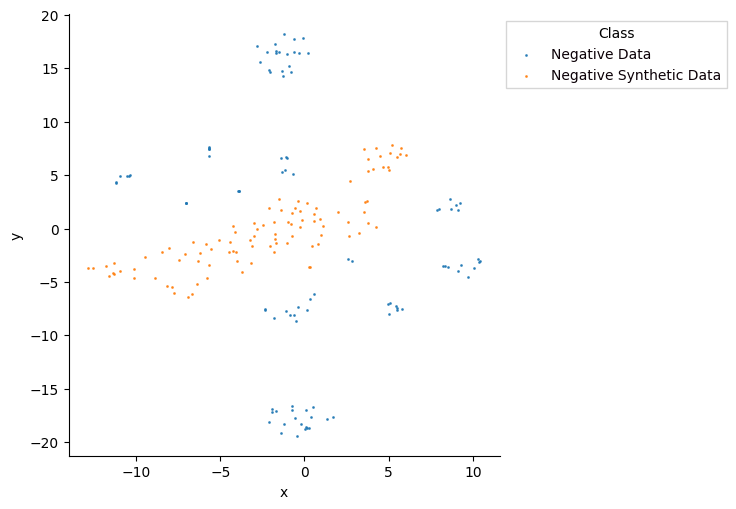}
    \subcaption{Epoch = 75}
  \end{minipage}
  \hfill
  \begin{minipage}[b]{0.48\textwidth}
    \includegraphics[width=5cm, height=4cm]{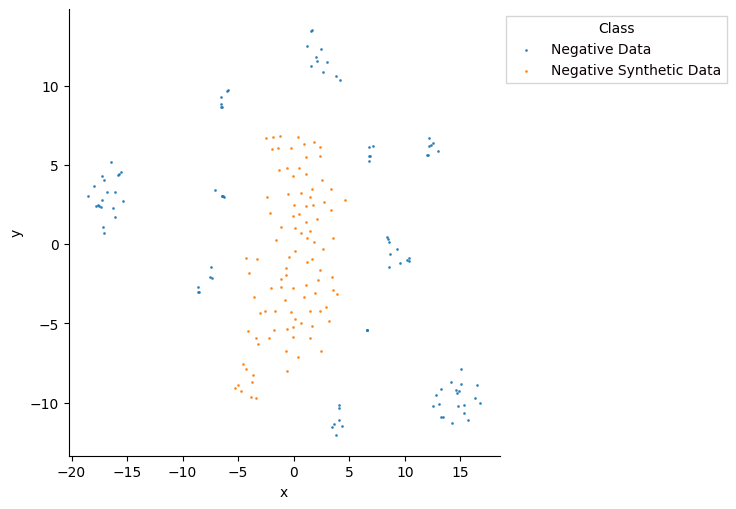}
    \subcaption{Epoch = 100}
  \end{minipage}
  \caption{t-SNE Plot of original and synthetic data for negative class}
  \label{fig4}
\end{figure}

\begin{figure}[H]
  \centering
  \begin{minipage}[b]{0.48\textwidth}
    \includegraphics[width=5cm, height=4cm]{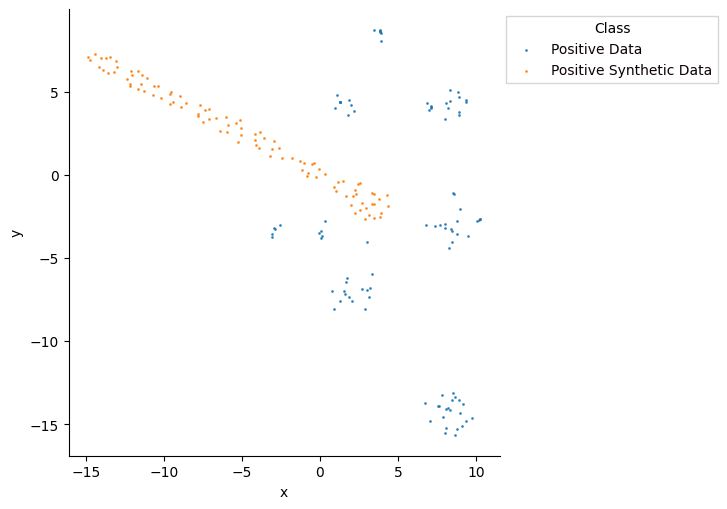}
    \subcaption{Epoch = 25}
  \end{minipage}
  \hfill
  \begin{minipage}[b]{0.48\textwidth}
    \includegraphics[width=5cm, height=4cm]{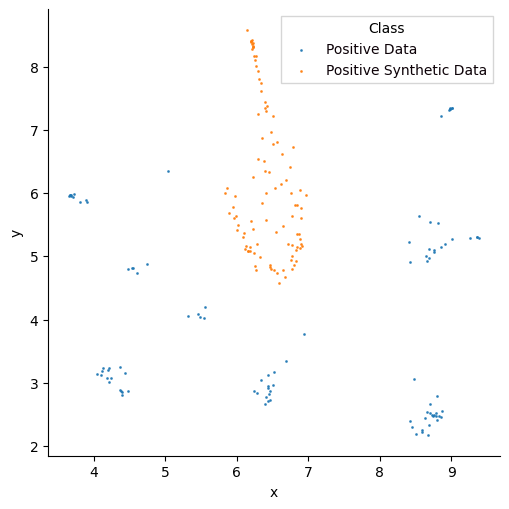}
    \subcaption{Epoch = 50}
  \end{minipage}
    \begin{minipage}[b]{0.48\textwidth}
    \includegraphics[width=5cm, height=4cm]{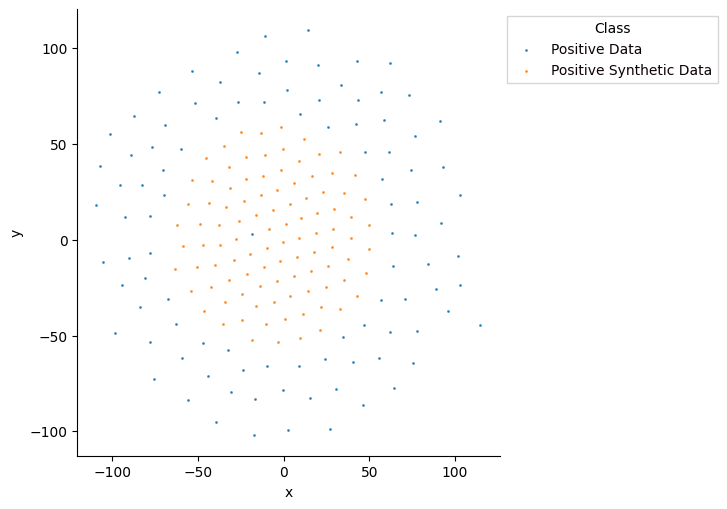}
    \subcaption{Epoch = 75}
  \end{minipage}
  \hfill
  \begin{minipage}[b]{0.48\textwidth}
    \includegraphics[width=5cm, height=4cm]{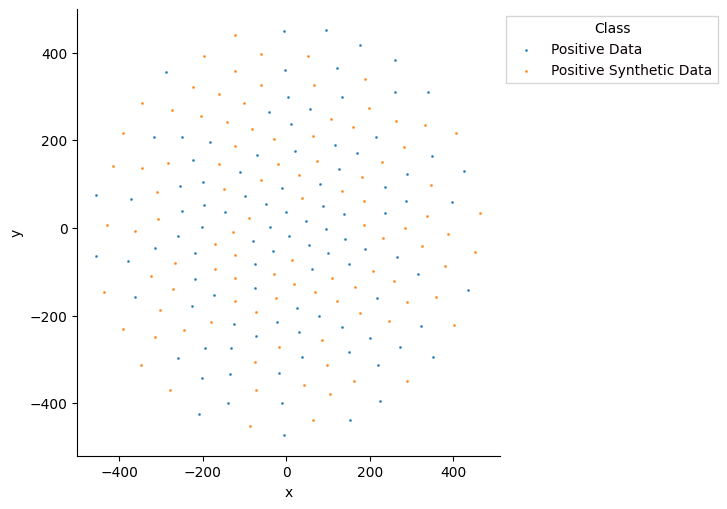}
    \subcaption{Epoch = 100}
  \end{minipage}
  \caption{t-SNE Plot of origianl and synthetic data for postive class}
  \label{fig5}
\end{figure}

\begin{figure}[H]
  \centering
  \begin{minipage}[b]{0.32\textwidth}
    \includegraphics[width=4.6cm, height=4cm]{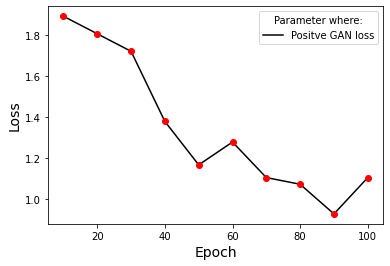}
    \caption{Postive loss}
    \label{fig6}
  \end{minipage}
  \hfill
  \begin{minipage}[b]{0.32\textwidth}
    \includegraphics[width=4.6cm, height=4cm]{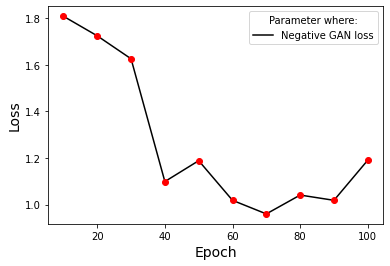}
    \caption{Negative loss}
    \label{fig7}
  \end{minipage}
  \hfill
  \begin{minipage}[b]{0.32\textwidth}
    \includegraphics[width=4.6cm, height=4cm]{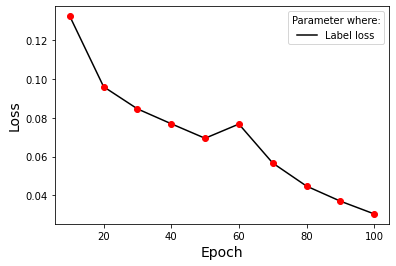}
    \caption{Label loss}
    \label{fig8}
  \end{minipage}
\end{figure}

The proposed GAN based model starts with some random values and tries to generate synthetic data, which helps to achieve a better F1 score. In the training process, after every epoch, we have calculated the F1 score for both the test datasets and observed a gradual improvement of the F1 score. The gradual change of precision, recall, and F1 score for the FEVER 1.0 and FEVER 2.0 is presented in  Fig. \ref{fig2} and Fig.  \ref{fig3}. Moreover, to visualize the distribution of original and synthetically data, the t-SNE plot of the positive and negative generated data is shown in Fig. \ref{fig4} and Fig. \ref{fig5}. The perplexity of the t-SNE plot is 30, and the learning rate is 120. It can be observed that the distribution of synthetically generated positive data is very similar to that of original positive text data, while the distribution of the negative synthetic data is similar to the original negative text data. The positive synthetic data is much more similar to the positive text data compared to the similarity between negative synthetic data and negative text data.

\begin{figure}[H]
  \centering
  \begin{minipage}[b]{0.32\textwidth}
    \includegraphics[width=4.6cm, height=4cm]{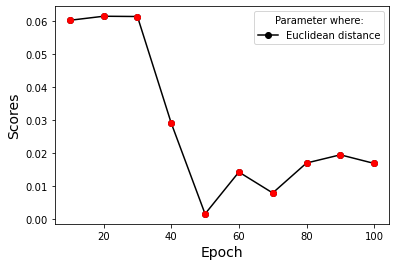}
    \subcaption{Cosine distance}
    \label{fig9}
  \end{minipage}
  \hfill
  \begin{minipage}[b]{0.32\textwidth}
    \includegraphics[width=4.6cm, height=4cm]{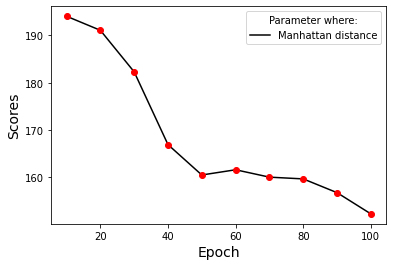}
    \subcaption{Manhattan distance}
    \label{fig10}
  \end{minipage}
  \hfill
  \begin{minipage}[b]{0.32\textwidth}
    \includegraphics[width=4.6cm, height=4cm]{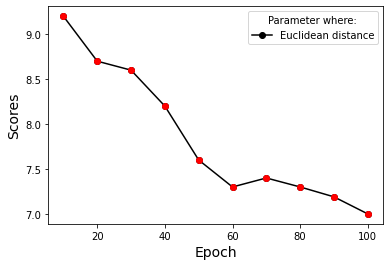}
    \subcaption{Euclidean distance}
    \label{fig11}
  \end{minipage}
  \caption{Similarity scores for positive data}
  \label{fig12}
\end{figure}

\begin{figure}[H]
  \centering
  \begin{minipage}[b]{0.32\textwidth}
    \includegraphics[width=4.6cm, height=4cm]{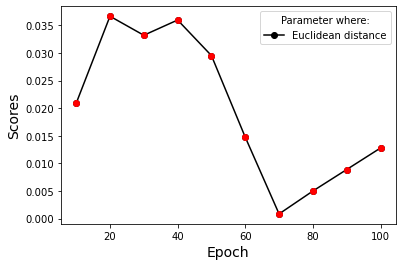}
    \subcaption{Cosine distance}
    \label{fig13}
  \end{minipage}
  \hfill
  \begin{minipage}[b]{0.32\textwidth}
    \includegraphics[width=4.6cm, height=4cm]{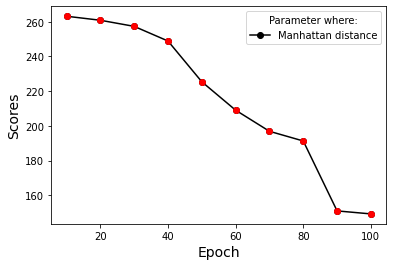}
    \subcaption{Manhattan distance}
    \label{fig14}
  \end{minipage}
  \hfill
  \begin{minipage}[b]{0.32\textwidth}
    \includegraphics[width=4.6cm, height=4cm]{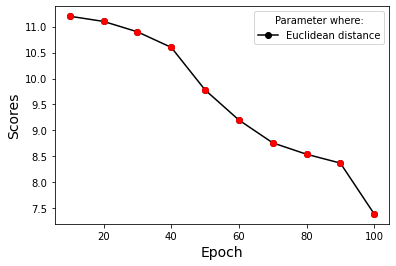}
    \subcaption{Euclidean distance}
    \label{fig15}
  \end{minipage}
  \caption{Similarity scores for negative data}
  \label{fig16}
\end{figure}

Fig. \ref{fig6}, \ref{fig7}, \ref{fig8} depicting the   positive loss, negative loss and label generating loss. We can see the  three losses are decreasing over epochs gradually, which also suggests that all the generator discriminator pairs are training to achieve the equilibrium state. To test the gradual progression of the synthetically generated data, we also measure the similarity scores between original (positive and negative) data and synthetic data (positive and negative) while training the model. It has been observed that for the generated data, the similarity score gradually improves over epochs, as shown in Fig. \ref{fig12} and \ref{fig16}. To measure the similarity 20,000  synthetically generated data are randomly selected and  Cosine similarity \cite{singhal2001modern}, Manhattan distance \cite{sinwar2014study}, Euclidean distance \cite{aggarwal2001surprising} are calculated.

\section{Conclusion}

This research proposes a multiple GAN-based model that employs the GAN's synthetic data generation capability to solve claim verification problems. The model generates synthetic data for supported, refuted claims and their class labels using three separate generator discriminator pairs. The synthetic data eventually helps in the fact-checking task for FEVER 1.0 and FEVER 2.0 test datasets.  The results have shown that the proposed model starts with random data generation, and as the training progresses, it generates synthetic data similar to the original data. Different statistical and analytical similarity metrics confirm that the similarity between original data and synthetically generated data increases as the training progresses. This gradual improvement of data quality shows the effectiveness of the model. The proposed model produces an F1 score of 0.65 $\pm$ 0.018 and 0.65 $\pm$ 0.051 for FEVER 1.0 and FEVER 2.0, respectively.  In the future, this model can be extended to a multi-class classifier, and a similar set of experiments can be carried out on other publicly available standard datasets to test this proposed model's effectiveness.



\bibliography{oasics-v2021-sample-article}

\begin{thebibliography}{10}

\bibitem{aggarwal2001surprising}
Charu~C Aggarwal, Alexander Hinneburg, and Daniel~A Keim.
\newblock On the surprising behavior of distance metrics in high dimensional
  space.
\newblock pages 420--434, 2001.

\bibitem{2}
Ramy Baly, Georgi Karadzhov, Dimitar Alexandrov, James Glass, and Preslav
  Nakov.
\newblock Predicting factuality of reporting and bias of news media sources.
\newblock {\em arXiv preprint arXiv:1810.01765}, 2018.

\bibitem{bert}
Jacob Devlin, Ming-Wei Chang, Kenton Lee, and Kristina Toutanova.
\newblock Bert: Pre-training of deep bidirectional transformers for language
  understanding.
\newblock {\em arXiv preprint arXiv:1810.04805}, 2018.

\bibitem{donahue2018adversarial}
David Donahue and Anna Rumshisky.
\newblock Adversarial text generation without reinforcement learning.
\newblock {\em arXiv preprint arXiv:1810.06640}, 2018.

\bibitem{svm}
Harris Drucker, Christopher~J Burges, Linda Kaufman, Alex Smola, and Vladimir
  Vapnik.
\newblock Support vector regression machines.
\newblock {\em Advances in neural information processing systems}, 9:155--161,
  1996.

\bibitem{friedman2002stochastic}
Jerome~H Friedman.
\newblock Stochastic gradient boosting.
\newblock {\em Computational statistics \& data analysis}, 38(4):367--378,
  2002.

\bibitem{fuglede2004jensen}
Bent Fuglede and Flemming Topsoe.
\newblock Jensen-shannon divergence and hilbert space embedding.
\newblock page~31, 2004.

\bibitem{Gardner2017AllenNLP}
Matt Gardner, Joel Grus, Mark Neumann, Oyvind Tafjord, Pradeep Dasigi,
  Nelson~F. Liu, Matthew Peters, Michael Schmitz, and Luke~S. Zettlemoyer.
\newblock Allennlp: A deep semantic natural language processing platform.
\newblock 2017.
\newblock \href {http://arxiv.org/abs/arXiv:1803.07640}
  {\path{arXiv:arXiv:1803.07640}}.

\bibitem{ge2013multi}
Liang Ge, Jing Gao, Xiaoyi Li, and Aidong Zhang.
\newblock Multi-source deep learning for information trustworthiness
  estimation.
\newblock pages 766--774, 2013.

\bibitem{gan}
Ian Goodfellow, Jean Pouget-Abadie, Mehdi Mirza, Bing Xu, David Warde-Farley,
  Sherjil Ozair, Aaron Courville, and Yoshua Bengio.
\newblock Generative adversarial nets.
\newblock {\em Advances in neural information processing systems},
  27:2672--2680, 2014.

\bibitem{leakgan}
Jiaxian Guo, Sidi Lu, Han Cai, Weinan Zhang, Yong Yu, and Jun Wang.
\newblock Long text generation via adversarial training with leaked
  information.
\newblock {\em arXiv preprint arXiv:1709.08624}, 2017.

\bibitem{lstm}
Sepp Hochreiter and J{\"u}rgen Schmidhuber.
\newblock Long short-term memory.
\newblock {\em Neural computation}, 9(8):1735--1780, 1997.

\bibitem{hou2017generative}
Ming Hou, Brahim Chaib-Draa, Chao Li, and Qibin Zhao.
\newblock Generative adversarial positive-unlabelled learning.
\newblock {\em arXiv preprint arXiv:1711.08054}, 2017.

\bibitem{krippendorff2018content}
Klaus Krippendorff.
\newblock Content analysis: An introduction to its methodology.
\newblock 2018.

\bibitem{cnn}
Steve Lawrence, C~Lee Giles, Ah~Chung Tsoi, and Andrew~D Back.
\newblock Face recognition: A convolutional neural-network approach.
\newblock {\em IEEE transactions on neural networks}, 8(1):98--113, 1997.

\bibitem{naive}
David~D Lewis.
\newblock Naive (bayes) at forty: The independence assumption in information
  retrieval.
\newblock pages 4--15, 1998.

\bibitem{li2014resolving}
Qi~Li, Yaliang Li, Jing Gao, Bo~Zhao, Wei Fan, and Jiawei Han.
\newblock Resolving conflicts in heterogeneous data by truth discovery and
  source reliability estimation.
\newblock pages 1187--1198, 2014.

\bibitem{mihalcea2009lie}
Rada Mihalcea and Carlo Strapparava.
\newblock The lie detector: Explorations in the automatic recognition of
  deceptive language.
\newblock pages 309--312, 2009.

\bibitem{neuendorf2015content}
Kimberly~A Neuendorf and Anup Kumar.
\newblock Content analysis.
\newblock {\em The international encyclopedia of political communication},
  pages 1--10, 2015.

\bibitem{random}
Mahesh Pal.
\newblock Random forest classifier for remote sensing classification.
\newblock {\em International journal of remote sensing}, 26(1):217--222, 2005.

\bibitem{pasternack2011making}
Jeff Pasternack and Dan Roth.
\newblock Making better informed trust decisions with generalized fact-finding.
\newblock 2011.

\bibitem{NEURIPS2019_9015}
Adam Paszke, Sam Gross, Francisco Massa, Adam Lerer, James Bradbury, Gregory
  Chanan, Trevor Killeen, Zeming Lin, Natalia Gimelshein, Luca Antiga, Alban
  Desmaison, Andreas Kopf, Edward Yang, Zachary DeVito, Martin Raison, Alykhan
  Tejani, Sasank Chilamkurthy, Benoit Steiner, Lu~Fang, Junjie Bai, and Soumith
  Chintala.
\newblock Pytorch: An imperative style, high-performance deep learning library.
\newblock pages 8024--8035, 2019.
\newblock URL:
  \url{http://papers.neurips.cc/paper/9015-pytorch-an-imperative-style-high-performance-deep-learning-library.pdf}.

\bibitem{scikit-learn}
F.~Pedregosa, G.~Varoquaux, A.~Gramfort, V.~Michel, B.~Thirion, O.~Grisel,
  M.~Blondel, P.~Prettenhofer, R.~Weiss, V.~Dubourg, J.~Vanderplas, A.~Passos,
  D.~Cournapeau, M.~Brucher, M.~Perrot, and E.~Duchesnay.
\newblock Scikit-learn: Machine learning in {P}ython.
\newblock {\em Journal of Machine Learning Research}, 12:2825--2830, 2011.

\bibitem{pennebaker2001linguistic}
James~W Pennebaker, Martha~E Francis, and Roger~J Booth.
\newblock Linguistic inquiry and word count: Liwc 2001.
\newblock {\em Mahway: Lawrence Erlbaum Associates}, 71(2001):2001, 2001.

\bibitem{pennington2014glove}
Jeffrey Pennington, Richard Socher, and Christopher~D Manning.
\newblock Glove: Global vectors for word representation.
\newblock pages 1532--1543, 2014.

\bibitem{3}
Ver{\'o}nica P{\'e}rez-Rosas, Bennett Kleinberg, Alexandra Lefevre, and Rada
  Mihalcea.
\newblock Automatic detection of fake news.
\newblock {\em arXiv preprint arXiv:1708.07104}, 2017.

\bibitem{7}
Ravali Pochampally, Anish Das~Sarma, Xin~Luna Dong, Alexandra Meliou, and
  Divesh Srivastava.
\newblock Fusing data with correlations.
\newblock pages 433--444, 2014.

\bibitem{popat2018declare}
Kashyap Popat, Subhabrata Mukherjee, Andrew Yates, and Gerhard Weikum.
\newblock Declare: Debunking fake news and false claims using evidence-aware
  deep learning.
\newblock {\em arXiv preprint arXiv:1809.06416}, 2018.

\bibitem{1}
Hannah Rashkin, Eunsol Choi, Jin~Yea Jang, Svitlana Volkova, and Yejin Choi.
\newblock Truth of varying shades: Analyzing language in fake news and
  political fact-checking.
\newblock pages 2931--2937, 2017.

\bibitem{rubin2015deception}
Victoria~L Rubin, Yimin Chen, and Nadia~K Conroy.
\newblock Deception detection for news: three types of fakes.
\newblock {\em Proceedings of the Association for Information Science and
  Technology}, 52(1):1--4, 2015.

\bibitem{scarselli2008graph}
Franco Scarselli, Marco Gori, Ah~Chung Tsoi, Markus Hagenbuchner, and Gabriele
  Monfardini.
\newblock The graph neural network model.
\newblock {\em IEEE Transactions on Neural Networks}, 20(1):61--80, 2008.

\bibitem{singhal2001modern}
Amit Singhal et~al.
\newblock Modern information retrieval: A brief overview.
\newblock {\em IEEE Data Eng. Bull.}, 24(4):35--43, 2001.

\bibitem{sinwar2014study}
Deepak Sinwar and Rahul Kaushik.
\newblock Study of euclidean and manhattan distance metrics using simple
  k-means clustering.
\newblock {\em Int. J. Res. Appl. Sci. Eng. Technol}, 2(5):270--274, 2014.

\bibitem{thorne2019adversarial}
James Thorne and Andreas Vlachos.
\newblock Adversarial attacks against fact extraction and verification.
\newblock {\em arXiv preprint arXiv:1903.05543}, 2019.

\bibitem{fever}
James Thorne, Andreas Vlachos, Christos Christodoulopoulos, and Arpit Mittal.
\newblock Fever: a large-scale dataset for fact extraction and verification.
\newblock {\em arXiv preprint arXiv:1803.05355}, 2018.

\bibitem{thorne2018fact}
James Thorne, Andreas Vlachos, Oana Cocarascu, Christos Christodoulopoulos, and
  Arpit Mittal.
\newblock The fact extraction and verification (fever) shared task.
\newblock {\em arXiv preprint arXiv:1811.10971}, 2018.

\bibitem{thorne2019fever2}
James Thorne, Andreas Vlachos, Oana Cocarascu, Christos Christodoulopoulos, and
  Arpit Mittal.
\newblock The fever2. 0 shared task.
\newblock pages 1--6, 2019.

\bibitem{wan2016truth}
Mengting Wan, Xiangyu Chen, Lance Kaplan, Jiawei Han, Jing Gao, and Bo~Zhao.
\newblock From truth discovery to trustworthy opinion discovery: An
  uncertainty-aware quantitative modeling approach.
\newblock pages 1885--1894, 2016.

\bibitem{fan}
Fan Yang, Eduard Dragut, and Arjun Mukherjee.
\newblock Claim verification under positive unlabeled learning.
\newblock {\em IEEE/ACM International Conference on Advances in Social Networks
  Analysis and Mining (ASONAM)}, 2020.

\bibitem{yang2018ti}
Yang Yang, Lei Zheng, Jiawei Zhang, Qingcai Cui, Zhoujun Li, and Philip~S Yu.
\newblock Ti-cnn: Convolutional neural networks for fake news detection.
\newblock {\em arXiv preprint arXiv:1806.00749}, 2018.

\end{thebibliography}

\appendix
\section{Mathematical Calculations}
\label{Appendix:A}
In the proposed methodology three GAN units are used. These three units are responsible for generating positive samples Equation \ref{Eq:1a}, negative samples Equation \ref{Eq:2a} and class labels Equation \ref{Eq:3a}. 

\begin{equation}
\label{Eq:1a}
\min_{G_p}\max_{D_p} V(D,G) = \mathop{\mathbb{E}_{x \sim p_p{(x)}}} log(D_p(x))+\mathop{\mathbb{E}_{z \sim p_z{(z)}}} log(1-D_p(G_p(z)))
\end{equation} 

\begin{equation}
\label{Eq:2a}
\min_{G_n}\max_{D_n} V(D,G) = \mathop{\mathbb{E}_{x \sim p_n{(x)}}} log(D_n(x))+\mathop{\mathbb{E}_{z \sim p_z{(z)}}} log(1-D_n(G_n(z)))
\end{equation} 

\begin{equation}
\label{Eq:3a}
\begin{split}
\min_{G_p,G_n,G_y}\max_{D_y} V(D,G) = \mathop{\mathbb{E}_{x \sim p{(x)}}} log(D_y(x))+ 
\pi_p\mathop{\mathbb{E}_{z \sim p_z{(z)}}} log(1-D_y(G_y(G_p(z))))+
\\ \pi_n\mathop{\mathbb{E}_{z \sim p_z{(z)}}} log(1-D_y(G_y(G_n(z))))
\end{split}
\end{equation}

To find the equilibrium condition, first, we need to find the optimal conditions for discriminators. Using the optimal conditions of the discriminators, the minimization conditions for the generator can be obtained. Considering the generators ($G_p$, $G_n$, $G_y$) are fixed, and $\pi_p$ and $\pi_n$ are the probabilities of positive and negative claims in the dataset. So in the in equilibrium condition the distribution of positive generated data ($p_{gp}(x)$) and negative generated data ($p_{gn}(x)$) will follow the below mentioned Equations \ref{Eq:5a} and \ref{Eq:6a}. 

\begin{equation}
\label{Eq:5a}
p_{gp}(x) = p_p(x)
\end{equation} 
\begin{equation}
\label{Eq:6a}
p_{gn}(x) = p_n(x)
\end{equation}

To find the optimal discriminator functions $D_p^{*}(x)$, $D_n^{*}(x)$, $D_y^{*}(x)$ we need to differentiate Equation \ref{Eq:1a}, \ref{Eq:2a} and \ref{Eq:3a}.

Let, $a = \mathop{\mathbb{E}_{x \sim p_p{(x)}}}$, 
$b = \mathop{\mathbb{E}_{z \sim p_p{(z)}}}$ and $t = D_p(x)$. Substitution $a$, $b$ and $x$ in Equation \ref{Eq:1a} we get Equation \ref{Eq:7a}.

\begin{equation}
\label{Eq:7a}
    l = a * log(t) + b * log(1-t) 
\end{equation}

Differentiating Equation \ref{Eq:7a} with respect to $t$ and equating the result with zero we can get the optimum value of $t$.

\begin{equation}
\label{Eq:8a}
     \diff{l}{t} = \frac{a}{t} - \frac{b}{(1-t)} = 0
\end{equation}

\begin{equation}
\label{Eq:9a}
     \frac{a}{t} = \frac{b}{(1-t)}
\end{equation}

\begin{equation}
\label{Eq:10a}
     t = \frac{a}{(a+b)}
\end{equation}

Therefore, substitution values of $a$ and $b$ the optimal discriminator function will be Equation \ref{Eq:11a}. The value of $b$ is $p_{gp}$ for the optimum condition as discussed earlier. 
\begin{equation}
\label{Eq:11a}
    D^{*}_P(x) = \frac{p_p(x)}{p_p(x)+p_{gp}(x)}
\end{equation}

Similarly, we can derive the equation for the negative discriminator as shown in Equation \ref{Eq:12a}.

\begin{equation}
\label{Eq:12a}
    D^{*}_n(x) = \frac{p_n(x)}{p_n(x)+p_{gn}(x)}
\end{equation}

To find the optimal values discriminator function for $D_y$ we need to differentiate Equation \ref{Eq:3a}. 

Let, $a = \mathop{\mathbb{E}_{x \sim p_p{(x)}}}$, 
$b = \pi_p*\mathop{\mathbb{E}_{z \sim p_p{(z)}}}$, 
$c = \pi_n*\mathop{\mathbb{E}_{z \sim p_p{(z)}}}$ and $t = D_y(x)$. Substitution $a$, $b$, $c$ and $x$ in Equation \ref{Eq:3a} we get Equation \ref{Eq:13a}.

\begin{equation}
\label{Eq:13a}
    l = a * log(t) + b * log(1-t) + c * log(1-t)
\end{equation}

Differentiating Equation \ref{Eq:13a} with respect to $t$ and equating the result with zero we can get the optimum value of $t$.

\begin{equation}
\label{Eq:14}
    \diff{l}{t} = \frac{a}{t} + \frac{b}{1-t} + \frac{c}{1-t} = 0
\end{equation}

\begin{equation}
\label{Eq:15a}
    t = \frac{a}{a+b+c} 
\end{equation}

Substituting the values of $a, b, c$ gives the Equation \ref{Eq:16a}. 

\begin{equation}
\label{Eq:16a}
\begin{split}
    \min_{G_p,G_n,G_y}\max_{D_y} V(D^*,G) = log\left(\frac{p(x)}{p(x)+\pi_pp_{gp}(x)+\pi_np_{gn}(x)}\right) + \\ \pi_plog\left(\frac{\pi_pp_{gp}(x)+\pi_np_{gn}(x)}{p(x)+\pi_pp_{gp}(x)+\pi_np_{gn}(x)}\right) + \\
    \pi_nlog\left(\frac{\pi_pp_{gp}(x)+\pi_np_{gn}(x)}{p(x)+\pi_pp_{gp}(x)+\pi_np_{gn}(x)}\right)
\end{split}
\end{equation}

Using Jensen–Shannon divergence (JSD) \cite{fuglede2004jensen}, we can show that the minimum value of the generators can be achieved when following conditions will be satisfied: 

\begin{equation}
\label{Eq:17a}
p_p(x) = p_{gp}(x)    
\end{equation}

\begin{equation}
\label{Eq:18a}
p_n(x) = p_{gn}(x)    
\end{equation}

\begin{equation}
\label{Eq:19a}
p_y(x) = \pi_pp_{gp}(x)+\pi_np_{gn}(x)
\end{equation}
\end{document}